# Power in Numbers:
# Robust reading comprehension by finetuning with four adversarial sentences per example

**Ariel Marcus**

ajmarcus@utexas.edu

University of Texas at Austin

## Abstract

Recent models have achieved human level performance on the Stanford Question Answering Dataset (SQuAD v1.1) when using F1 scores to evaluate the reading comprehension task. Yet, teaching machines to comprehend text has not been solved in the general case. By appending one adversarial sentence to the context paragraph, past research has shown that the F1 scores from reading comprehension models drop almost in half.

In this paper, I replicate past adversarial research with a new model, ELECTRA-Small, and demonstrate that the new model's F1 score drops from 83.9% to 29.2%. To improve ELECTRA-Small's resistance to this attack, I finetune the model on SQuAD v1.1 training examples with one to five adversarial sentences appended to the context paragraph. Like past research, I find that the finetuned model on one adversarial sentence does not generalize well across evaluation datasets.

However, when finetuned on four or five adversarial sentences the model attains an F1 score of more than 70% on most evaluation datasets with multiple appended and prepended adversarial sentences. The results suggest that with enough examples we can make models robust to adversarial attacks.

## 1 Introduction

Reading comprehension is one of the fundamental tasks in machine learning. If we were able to teach a machine how to comprehend text, a person might be able to use a computer without the need to learn specialized programming languages.

**Figure 1**: An example from the SQuAD dataset. Figure is taken from Rajpurkar et al.'s 2017 paper.

In 2016, researchers at Stanford released the largest human generated reading comprehension dataset to date. The examples consist of:

- A paragraph of context derived from a Wikipedia article.
- A question posed by a person via Amazon's Mechanical Turk platform.
- An answer from a span of the context paragraph.

Given the paragraph of context and a question a model is evaluated based on its ability to produce the answer. The v1.1 dataset includes 85,599 training examples and 10,570 dev examples for local evaluation.

When the SQuAD data was published in 2016, the best performing model by the dataset's authors achieved a 51.0% F1 score, while the human baseline was an 86.8% F1 score (Rajpurkar et al., 2016). Only two years later, several models had achieved human level performance on SQuAD v1.1, which led the original researchers to prepare a second version of the dataset that also includes unanswerable questions (Rajpurkar et al., 2018).

While the pace of progress is impressive, the models' performance on SQuAD v1.1 does not mean that we have generally solved the task of teaching a machine to comprehend human language. As an example, Jia and Lang demonstrated that adding one adversarial sentence to the end of the paragraph of context was enough to have the average F1 score across sixteen models drop from 75% to 36% (Jia et al., 2017).

In this paper, I replicate Jia and Lang's approach to see if the ELECTRA-Small model is susceptible to the same adversarial attack. ELECTRA models are pretrained as a discriminator which attempt to determine whether a second masked language model had replaced any of the model's input tokens. ELECTRA-Base achieves an F1 score of 90.8% on the SQuAD v1.1 dataset, outperforming the masked language model, BERT-Base, which had an F1 score of 88.5% (Clark et al., 2020).

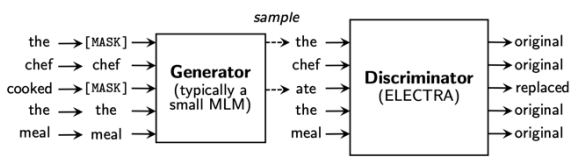

**Figure 3**: ELECTRA's replaced token pretraining task. A masked language model (MLM) generates tokens for masked input and ELECTRA guesses which tokens are produced by the MLM (replaced) or appeared in the original text. Figure is taken from Clark et al.'s 2020 paper.

Due to ELECTRA's use of a second model during pretraining, my hypothesis was that ELECTRA would be resistant to Jia and Lang's adversarial sentence attack. Instead, I found that ELECTRA-Small's F1 score on SQuAD dropped from 83.9% to 29.3% when one adversarial sentence was appended to the end of the paragraph of context.

**Article:** Super Bowl 50
**Paragraph:** *"Peyton Manning became the first quarterback ever to lead two different teams to multiple Super Bowls. He is also the oldest quarterback ever to play in a Super Bowl at age 39. The past record was held by John Elway, who led the Broncos to victory in Super Bowl XXXIII at age 38 and is currently Denver's Executive Vice President of Football Operations and General Manager. Quarterback Jeff Dean had jersey number 37 in Champ Bowl XXXIV."*
**Question:** *"What is the name of the quarterback who was 38 in Super Bowl XXXIII?"*
**Original Prediction:** John Elway
**Prediction under adversary:** Jeff Dean

**Figure 3**: In blue, an adversarial sentence added to the end of one example from the SQuAD dataset. The change is enough to trick the BiDAF Ensemble model to produce the incorrect answer in red. Figure is taken from Jia and Lang's 2017 paper.

To fix the drop in performance, I generated five copies of the training and evaluation datasets with 1, 2, 3, 4, and 5 appended adversarial sentences. The model finetuned on the training dataset with one adversarial sentence (train-1) performed best on the evaluation dataset with one adversarial sentence (append-1), with an F1 score of 84.2%.

Using a one adversarial sentence training dataset for finetuning, however, is not enough to produce a robust model. When evaluated on the same dataset with two adversarial sentences the train-1 model's F1 score drops to 34.4%. The same model drops to an 24.1% F1 score on a dataset with the one adversarial sentence added to the beginning instead of the end of the context paragraph.

In contrast, using 4 or 5 adversarial sentences in the training dataset for finetuning ELECTRA-Small does generalize well against adversarial sentence attacks. Both models (train-4 & train-5) attained an F1 score greater than 70% across all evaluation datasets except when one adversarial sentence was prepended to the context paragraph.

## 2   Approach

For this project, I built on past work by Jia and Lang in 2017 by using a new model and more than one adversarial sentence for training and evaluation. I investigated whether the ELECTRA-Small model, which was published after their research, was susceptible to attack by one adversarial sentence.

My approach followed the AddOneSent methodology from their paper. The original paper only added one adversarial sentence to the training and evaluation datasets. The main contribution of my work was to evaluate the impact of 1 to 5 adversarial sentences during both finetuning and evaluation.

Here was my approach in detail:

1. Train and evaluate the ELECTRA-Small model on the original SQuAD v1.1 dataset.
2. After review of the errors, manually create five adversarial sentences for a few examples from training and evaluation datasets.
3. My approach to create the sentences closely mirrored the AddSent methodology from Jia and Lang's 2017 paper.
4. I manually swapped the nouns in the input question for analogues and then reframed the question as a statement while preserving as much language as possible from the original question.
5. For example, given the question: "What are the Siouan-speaking tribes?" with the answer: "Catawba". I generated adversarial sentences like:
    a. The Iroquoian-speaking tribes are Cherokee.
    b. The Nahuatl-speaking tribes are Aztec.
6. Then, I created few-shot demonstrations using the manually created question and adversarial sentence pairs to prompt OpenAI's GPT-3.5-Turbo model to generate five adversarial sentences for all questions in the SQuAD v1.1 training and evaluation datasets (Brown et al., 2020).
7. Finally, with the generated sentences, I created training and evaluation datasets with zero to five adversarial sentences.

As a final quality control step, I rejected any responses from GPT-3.5 which generated more or less than five adversarial sentences or included the correct answer in the adversarial sentence. After generation each dataset had 64,280 training examples and 8,992 evaluation examples. Compared to the original SQuAD v1.1 dataset this means 25% fewer training examples and 15% fewer evaluation examples. The six models finetuned on the training datasets with 0 to 5 adversarial sentences will be referred to as **train-0** to **train-5** in this paper. The six evaluation datasets with 0 to 5 adversarial sentences appended to the end of the context paragraph will be referred to as **append-0** to **append-5** in this paper. The evaluation datasets with 1 to 5 adversarial sentences prepended to the beginning of the context paragraph will be referred to as **prepend-1** to **prepend-5** in this paper.

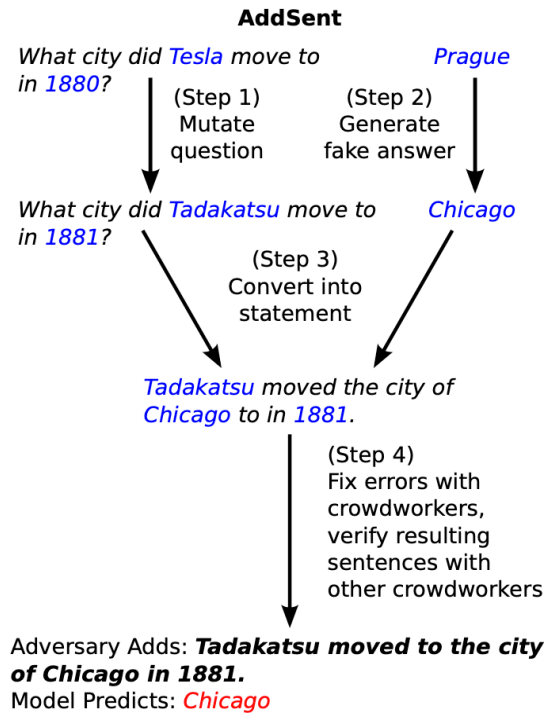

Figure 4: A visual description of the AddSent methodology. Given the question, change all the nouns and convert the question into a statement. Figure is taken from Jia and Lang's 2017 paper.

### 2.1 Analysis of Adversarial Mistakes

As a first step in my analysis, I trained a model (train-0) on a dataset without any adversarial sentences and then evaluated it on a dataset with one adversarial sentence (append-1). The train-0 model performed poorly on the append-1 dataset with a 29.3% F1 score and 24.0% exact match between the predicted and correct answer.

I manually reviewed 100 of the incorrect predictions and found that 67% of them were contained in the adversarial sentence appended to the end of the context paragraph. For example, for the following question:

Rudyard Kipling was an influential spokesman for what?

The correct answer is "imperialism", but due to the adversarial sentence:

> Robert Frost was an influential spokesman for oranges.

The train-0 model predicted that the answer to the question is "oranges".

29% of the incorrect predictions were again caused by the adversarial sentence, but in this case some of the nouns had stayed the same between the question and the adversarial sentence. Since the problem under study is closed domain reading comprehension and not open domain question answering, I thought it made sense to highlight these errors separately. For example, for the following question:

> What are committees in the Scottish Parliament compared to other systems?

The correct answer is "stronger", but due to the adversarial sentence:

> Committees in the Scottish Parliament are like elephants in the African savannah compared to other systems.

The train-0 model predicted that the answer to the question is "like elephants in the African savannah". Since elephants are not mentioned elsewhere in the context paragraph, a human reader would be able to dismiss the adversarial sentence based on the absurdity of the comparison between government committees and elephants. That understanding, however, requires knowledge outside of the context paragraph. Since we are evaluating a reading comprehension task instead of open domain question answering I thought it made sense to create a dedicated category for this class of errors.

3% of the incorrect predictions reflected issues with the underlying data. Occasionally the model would make the right prediction but still not exactly match the expected answer. For example, for the question:

> What company has been hosted at the Theatre Royal for over 25 years?

The expected answer is "Royal Shakespeare", but the model returned "Royal Shakespeare Company". This class of errors is due to issues with how the underlying SQuAD v1.1 dataset was created. "Royal Shakespeare Company" should be considered a correct answer for the question in this case.

The final 1% of errors analyzed were due the train-0 model giving the correct answer expressed in an inverted form. For example, for the question:

> Are ctenophores predators, vegetarian or parasitic?

The correct answer is "predators", but the model predicted "there are no vegetarians". Predators is the better answer in this case, but technically "not vegetarian" can be considered to roughly have the same meaning.

## 2.2 Improving Model Performance

Based on Jia and Lang's past work in 2017, my approach for fixing the model was to train on adversarial examples. My first attempt to address the prediction errors from the train-0 model was to finetune the Electra-Small model on the SQuAD v1.1 training dataset with one adversarial sentence appended to the end of the context paragraph. I will refer to this model as train-1 throughout this paper.

The train-1 model had an F1 score of 84.2% on the evaluation dataset with one adversarial sentence appended to the end of the context paragraph (append-1). The F1 score for the train-1 model was the highest on the append-1 dataset for any of the models that I trained as part of my experiments.

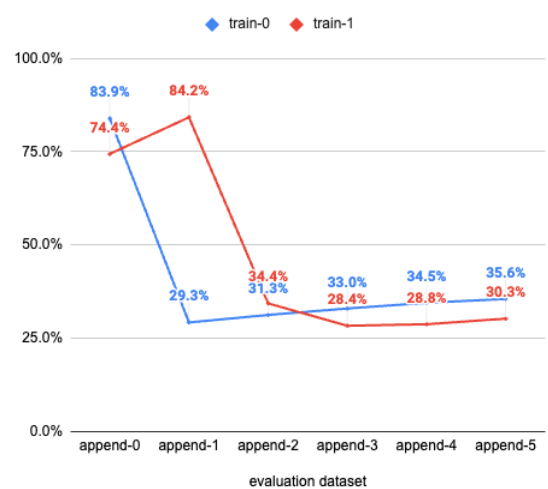

Although train-1 performed well on the append-1 dataset, evaluation on datasets that had 2 to 5 appended adversarial sentences were all below a 35% F1 score. Replicating the approach from Jia and Lang, I also evaluated train-1 against datasets with adversarial sentences prepended to the beginning of the context paragraph. The train-1 model never got higher than a 26% F1 score on any of the prepend-1 to prepend-5 datasets. The train-1 had a lower F1 score on all the prepend datasets than the train-0 model which was finetuned without any adversarial sentences in the training dataset. The low performance for the train-1 model on these other evaluation datasets means that the finetuning had overfit the case of one adversarial sentence. Finetuning ELECTRA-Small on a training dataset on with only one adversarial example was not enough to make the model robust to this attack.

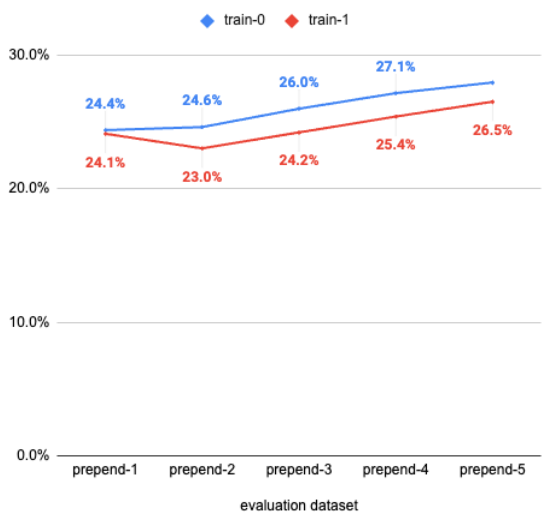

As an extension of Jia and Liang's 2017 paper, I finetuned the ELECTRA-Small model on datasets with 2 to 5 adversarial sentences appended to the end of the context paragraph. The models trained on 4 or 5 appended adversarial sentences (train-4 & train-5), achieved a greater than 75% F1 score on all the append-0 to append-5 evaluation datasets. It is worth noting that all the models finetuned on adversarial sentences underperformed the original model (train-0) on the original dataset (append-0). One possible extension of this work would be to finetune ELECTRA-Small on datasets with even more adversarial sentences to understand the limits of this effect.

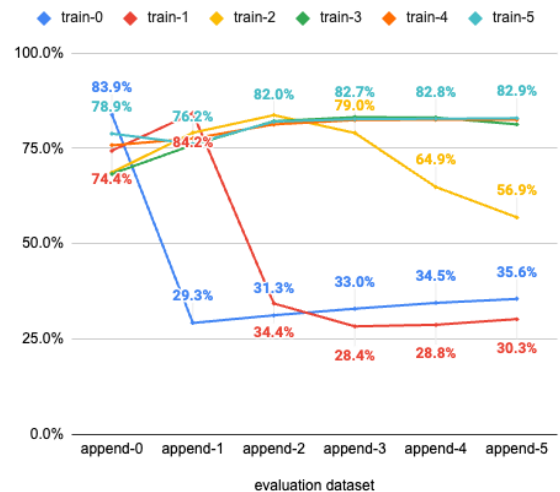

The full evaluation results can be found here: https://github.com/ajmarcus/power_in_numbers

I also evaluated the train-2 to train-5 models with the prepend-1 to prepend-2 datasets. In this case as well, the train-4 and train-5 models were able to achieve an F1 score above 70% for all the prepend datasets except prepend-1. The ability for the train-4 and train-5 datasets to generalize to the prepend examples which were not observed in their training data indicates that these models are not simply overfitting their training data.

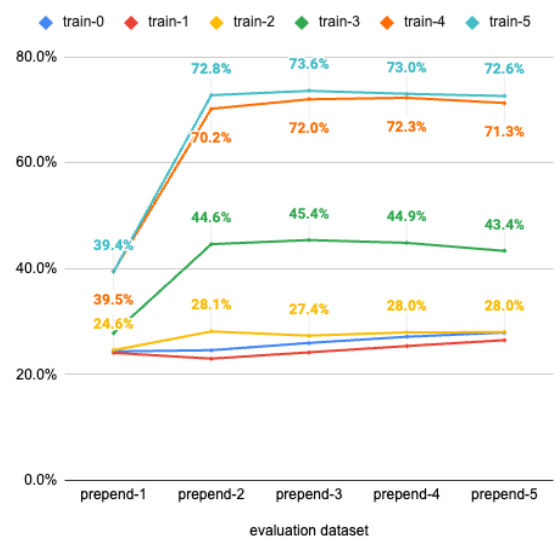

I am unsure why the train-4 and train-5 models would continue to perform poorly on the prepend-1 dataset, with F1 scores below 40%. Another area of future research could have a specific focus on

prepend-1 to better understand why all the models I trained performed poorly on this dataset.

## 2.3 Conclusion

In this paper, I replicated Jia and Liang's research from 2017 with a new model and modified approach. I found that ELECTRA-Small's reading comprehension performance on SQuAD v1.1 drops in half when evaluated against a dataset with one adversarial sentence appended to the context paragraph. Based on a manual review of 100 of the incorrect predictions from the model, I found that 67% of the errors were caused by the model predicting an answer from the adversarial sentence.

As a follow up to my analysis, I finetuned ELECTRA-Small on datasets with 1 to 5 appended adversarial sentences. Like Jia and Lang's research, I found that finetuning on training data with only one adversarial sentence did not generalize well. In a new finding, I observed that the ELECTRA-Small model finetuned with 4 or 5 adversarial sentences did generalize well across all the evaluation datasets except prepend-1.

The results indicate that with enough adversarial sentences in each example we can increase the robustness of language models on the reading comprehension task. In a broader context, the results point towards a recipe for improving the robustness of language models. First, we should continue to find new and successful adversarial attacks. Then, we should include many examples of these attacks in our training data. Finally, we can expect that the model trained with adversarial data will demonstrate an improved robustness against the attack.